%% file: main.tex
\documentclass{article}

\usepackage{microtype}
\usepackage{bbm}
\usepackage{float}
\usepackage{enumitem}
\usepackage{algcompatible}
\usepackage{graphicx}
\usepackage{soul}
\usepackage{subfigure}
\usepackage{booktabs} 

\usepackage{hyperref}

\usepackage[accepted]{misc/icml2025}

\usepackage{amsmath}
\usepackage{amssymb}
\usepackage{mathtools}
\usepackage{amsthm}

\usepackage[framemethod=TikZ]{mdframed}
\mdfsetup{
    middlelinecolor =   none,
    middlelinewidth =   1pt,
    backgroundcolor =   blue!5,
    roundcorner     =   5pt,
}

\usepackage[capitalize,noabbrev]{cleveref}

\theoremstyle{plain}

\theoremstyle{definition}

\theoremstyle{remark}

\usepackage[textsize=tiny]{todonotes}

\newcommand{\algo}{\texttt{PrefVLM}}

\icmltitlerunning{Minimizing Human Feedback in Reinforcement Learning using Vision-Language Models}

\begin{document}

\twocolumn[
\icmltitle{Preference VLM: Leveraging VLMs for Scalable Preference-Based Reinforcement Learning}



\icmlsetsymbol{equal}{*}

\begin{icmlauthorlist}
\icmlauthor{Udita Ghosh}{ucr}
\icmlauthor{Dripta S. Raychaudhuri}{aws}
\icmlauthor{Jiachen Li}{ucr}
\icmlauthor{Konstantinos Karydis}{ucr}
\icmlauthor{Amit Roy-Chowdhury}{ucr}
\end{icmlauthorlist}

\icmlaffiliation{ucr}{University of California, Riverside}
\icmlaffiliation{aws}{AWS AI Labs}

\icmlcorrespondingauthor{Udita Ghosh}{ughos002@ucr.edu}

\icmlkeywords{Vision-language, Reinforcement learning, Embodied AI}

\vskip 0.3in
]



\printAffiliationsAndNotice{\icmlEqualContribution} 

\input{sections/0_abstract}
\input{sections/1_introduction}

\input{sections/2_related}

\input{sections/3_method}
\input{sections/4_experiments}
\input{sections/5_conclusion}

\bibliography{references}
\bibliographystyle{misc/icml2025}

\input{sections/6_appendix}

\end{document}

%% file: sections/0_abstract.tex
\begin{abstract}
Preference-based reinforcement learning (RL) offers a promising approach for aligning policies with human intent but is often constrained by the high cost of human feedback. In this work, we introduce \algo, a framework that integrates Vision-Language Models (VLMs) with selective human feedback to significantly reduce annotation requirements while maintaining performance. Our method leverages VLMs to generate initial preference labels, which are then filtered to identify uncertain cases for targeted human annotation. Additionally, we adapt VLMs using a self-supervised inverse dynamics loss to improve alignment with evolving policies. Experiments on Meta-World manipulation tasks demonstrate that \algo~achieves comparable or superior success rates to state-of-the-art methods while using up to $2\times$ fewer human annotations. Furthermore, we show that adapted VLMs enable efficient knowledge transfer across tasks, further minimizing feedback needs. Our results highlight the potential of combining VLMs with selective human supervision to make preference-based RL more scalable and practical.

\end{abstract}

%% file: sections/1_introduction.tex
\section{Introduction} \label{sec:intro}

Deep reinforcement learning (RL) has achieved remarkable performance across a broad range of sequential decision-making tasks, from mastering games~\cite{mnih2015human,silver2017mastering,silver2018general} to solving complex robotic challenges~\cite{duan2016benchmarking,levine2016end,kaufmann2023champion}. Despite these successes, a significant obstacle to deploying RL in real-world settings lies in designing reward functions that consistently guide agents toward desired behaviors.
\begin{figure}[t]
    \centering
    \includegraphics[width=\columnwidth]{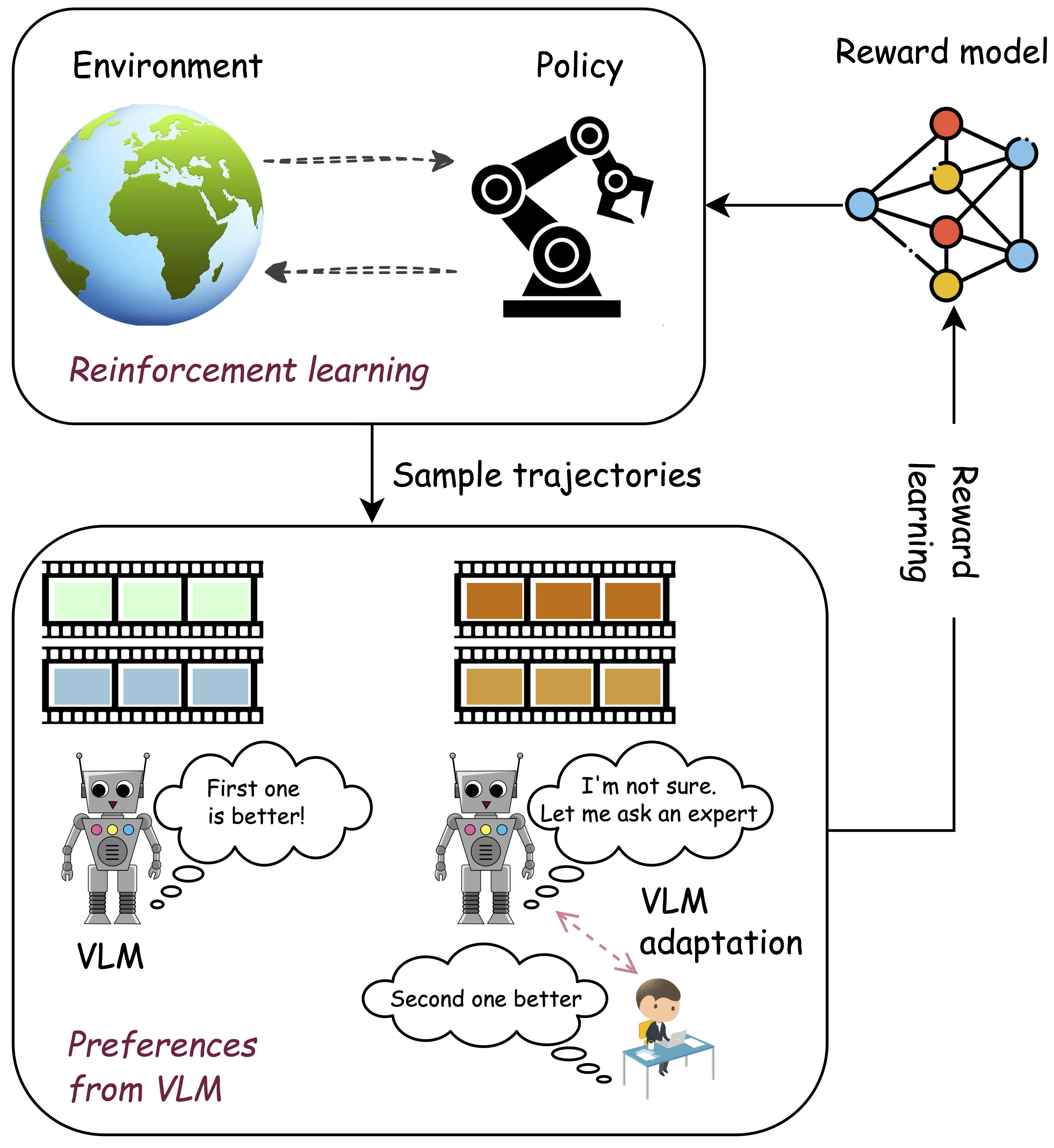}
    \caption{\algo~leverages a VLM to obtain preference labels over pairs of the agent's trajectory segments. These preference labels are then used to train a reward function. In scenarios where the VLM exhibits high uncertainty, \algo~can seamlessly incorporate human feedback to refine its understanding and adapt the VLM to the specific environment. By combining machine-generated and expert-guided feedback, \algo~learns high-quality reward function while significantly reducing the amount of human supervision required compared to existing preference-based RL methods.}
    \label{fig:teaser}
\end{figure}
Manually crafting reward functions is particularly challenging for complex, long-horizon tasks, as it often demands substantial human expertise and ad hoc engineering. Preference-based RL~\cite{christiano2017deep,lee2021pebble} has emerged as a promising alternative, bypassing the need for hand-crafted rewards by relying on human-provided preferences over pairs of agent behaviors. These preferences are then used to infer a reward function for training the agent. While preference-based RL simplifies reward design, it remains limited by its heavy reliance on human feedback, which can be resource-intensive and thus impede scalability. 

To address this limitation, recent efforts have explored the use of vision-language models (VLMs)~\cite{radford2021learning} as a scalable and intuitive means to specify rewards through natural language descriptions (\textit{e.g.}, ``open the door''). VLMs align visual observations with textual descriptions in a shared latent space, enabling dense, zero-shot reward generation across diverse tasks~\cite{mahmoudieh2022zero,adeniji2023language,ma2023liv,rocamonde2023vision,sontakke2024roboclip}. However, employing VLMs alone as reward functions introduces challenges: their outputs are often noisy and lack the granularity required for fine-grained tasks like robotic manipulation~\cite{fu2024furl}.

In this paper, we propose \algo, a novel framework that enhances the efficiency of preference-based RL by synergistically integrating VLMs with minimal human feedback. Unlike existing methods that rely exclusively on VLMs or extensive human supervision for reward generation, \algo~uses VLMs to generate coarse, trajectory-level preference signals, which are subsequently refined with targeted human input in cases of high uncertainty. This approach reduces the annotation burden ($\sim 2\times$) while ensuring the precision required for complex tasks. 
Figure~\ref{fig:teaser} illustrates our approach.

Our approach has two core components. 
\textit{First}, we introduce a parameter-efficient fine-tuning strategy for VLMs that combines an unsupervised dynamics-aware objective with sparse human feedback, improving the quality of preference labels. 
While prior efforts have sought to reduce noise in VLM rewards by incorporating expert trajectories and additional environment-based rewards~\cite{fu2024furl}, obtaining high-quality demonstrations remains a major challenge because of the high cost of data collection~\cite{akgun2012keyframe} and substantial domain gaps~\cite{smith2019avid}. 
In contrast, providing preference feedback is a more accessible and resource-efficient alternative~\cite{hejna2023few}. 
\textit{Second}, we leverage robust training techniques~\cite{song2022learning,cheng2024rime} to improve the learning process. 
By dynamically constraining the KL divergence between the reward model and VLM predictions, we identify and prioritize reliable samples for training while isolating uncertain ones for targeted human annotation. 
This two-pronged approach ensures both efficiency and precision in reward learning.

We evaluate \algo~on five robotic manipulation tasks and demonstrate that it produces high-quality reward functions for training optimal policies. 
Our results show that \algo~reduces human feedback requirements by \emph{half} compared to existing preference-based RL methods while maintaining comparable or superior performance. 
These findings highlight how VLMs can be leveraged to make preference-based RL significantly more efficient without compromising precision or scalability.

\begin{mdframed}[innertopmargin=0pt,leftmargin=0pt, rightmargin=0pt, innerleftmargin=10pt, innerrightmargin=10pt, skipbelow=0pt]
\paragraph{Our work has three main contributions:}
\begin{enumerate}[leftmargin=*]
    \item Conceptually, we demonstrate that VLMs are effective at providing coarse, trajectory-level preferences, making them a scalable and efficient alternative for reward model learning in preference-based RL. By leveraging these models, we show that the dependence on extensive human feedback can be significantly reduced.
    \item Methodologically, we propose a new approach, \algo, that leverages VLMs to generate preference labels over trajectory pairs and integrates them into a preference learning framework. Our method combines parameter-efficient fine-tuning of VLMs using minimal human feedback, and robust training techniques to handle noise in machine-generated preferences.
    \item Empirically, we evaluate \algo~on robotic manipulation tasks and show that it achieves comparable or superior performance to existing preference-based RL methods while reducing the need for human supervision by half, demonstrating its scalability and practicality.
\end{enumerate}
\end{mdframed}

%% file: sections/2_related.tex
\section{Related Works} \label{sec:related_works}

\noindent \textbf{Designing reward functions.} 
Designing effective reward functions remains a fundamental challenge in RL~\cite{singh2009rewards,sutton2018reinforcement}. Traditional methods often rely on manual trial-and-error processes, which require substantial domain expertise and struggle to scale to complex, long-horizon tasks~\cite{booth2023perils,knox2023reward}. Inverse reinforcement learning offers an alternative by attempting to infer reward functions from expert demonstrations~\cite{abbeel2004apprenticeship,ziebart2008maximum,ho2016generative}. Evolutionary algorithms have also been explored for automated reward function discovery~\cite{niekum2010genetic,chiang2019learning}. 

Recently, foundation models have emerged as a promising avenue for reward function generation. Multimodal Large language models (MLLMs) have been employed to derive reward functions directly from natural language task descriptions~\cite{yu2023language,ma2023eureka,wang2024}, although these approaches either often assume access to the environment's source code or are too expensive for repeated queries. VLMs present another promising direction~\cite{mahmoudieh2022zero,adeniji2023language,ma2023liv,rocamonde2023vision,sontakke2024roboclip}, leveraging task descriptions to generate rewards by aligning the agent's visual observations with the task description in the joint language-image embedding space. However, rewards generated by VLMs often exhibit noise and lack the granularity needed for precise tasks~\cite{fu2024furl}. To address these challenges, our work introduces a hybrid approach that synergistically combines the coarse preference signals provided by VLMs with targeted human feedback. Instead of relying on VLMs as zero-shot reward generators, we leverage their ability to infer trajectory-level preferences as a foundation, while strategically incorporating human annotations to refine these preferences in uncertain cases.

\noindent \textbf{Learning from human feedback.} 
Incorporating human feedback has been widely explored across domains, including natural language processing~\cite{ouyang2022training,rafailov2024direct} and robotics~\cite{lee2021pebble}. In RL, preference-based frameworks have proven particularly effective, with \citet{christiano2017deep} pioneering the use of human-provided trajectory comparisons to guide learning. Subsequent works, such as PEBBLE~\cite{lee2021pebble}, enhanced feedback efficiency by incorporating unsupervised exploration, while SURF~\cite{park2022surf} augmented preference datasets using semi-supervised learning.

These approaches are rooted in the observation that humans often find it easier to provide relative judgments, \emph{i.e.}, comparing behaviors as better or worse, rather than defining absolute metrics. 
In this work, we extend this idea by integrating VLMs to generate preferences over pairs of an agent's trajectories. 
This approach reduces the annotation cost of human feedback while maintaining performance.

\noindent \textbf{Learning from noisy labels.} 
The challenge of learning from noisy or imprecise labels has been extensively studied in supervised learning, particularly with the rise of machine-generated annotations~\cite{wang2022self}. Robust training methods address this issue through a variety of strategies, including architectural modifications~\cite{goldberger2017training}, regularization techniques~\cite{lukasik2020does}, specialized loss functions~\cite{zhang2018generalized}, and sample selection mechanisms~\cite{wang2021denoising}.

In our framework, the preference labels generated by VLMs may contain noise due to the inherent limitations of these models. To handle this, we adopt a sample selection strategy inspired by RIME~\cite{cheng2024rime}, which identifies confident labels for training while flagging uncertain samples for human refinement. This combination of robust training and targeted human feedback ensures that our reward model remains both accurate and efficient.

%% file: sections/3_method.tex
\section{Preliminaries} \label{sec:background}

We consider the standard RL setup~\cite{sutton2018reinforcement} where an agent interacts with an environment in discrete time. Formally, at each timestep $t$, the agent observes a state $s_t$ from the environment and selects an action $a_t$ according to its policy $\pi$. In response, the environment provides a reward $r_t$ and transitions the agent to the next state $s_{t+1}$. The return $R_t = \sum_{k=0}^\infty \gamma^k r_{t+k}$ represents the discounted sum of future rewards starting at timestep $t$, where $\gamma \in [0, 1)$ is the discount factor. The goal of RL is to maximize the expected return from each state $s_t$ under the agent's policy.

\begin{figure}[!t]
    \centering
    \includegraphics[width=0.48\textwidth]{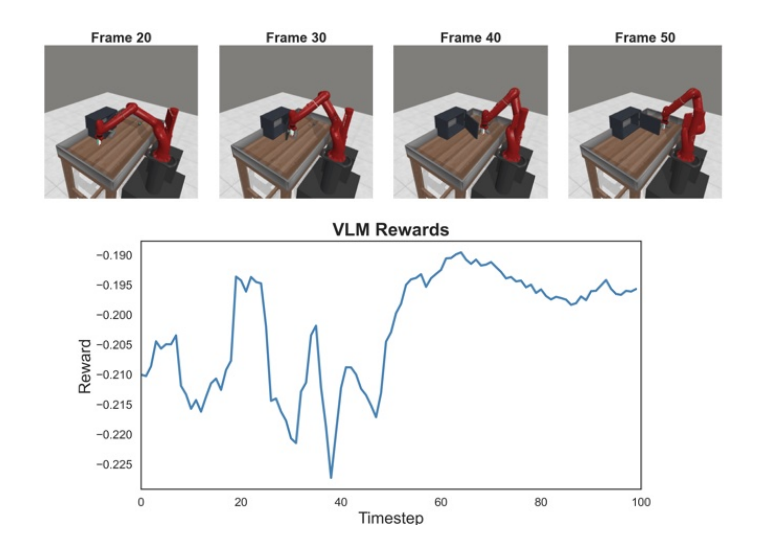}
    \caption{VLM reward (Eqn.~\ref{eq:cosine_sim}) for an optimal trajectory given the task description \textit{``Open a door with a revolving joint.''} Although the reward reflects partial task progression, it is noisy and poorly aligned with the actual task progress, as evident from the image observations. 
    }
    \label{fig:noisy_reward}
\end{figure}
\subsection{Learning rewards from preferences} \label{sec:pref}
We adopt the standard preference-based learning framework, where a teacher provides preferences over pairs of trajectory segments, and a reward function $r_\theta$ is learned to align with these preferences. Formally, a trajectory segment $\sigma$ is defined as a sequence of observations and actions: $\sigma = \{(s_1, a_1), (s_2, a_2),...,(s_T, a_T)\}$. Given a pair of segments $(\sigma_0, \sigma_1)$, preferences are expressed as $y \in \{(0,1), (1,0)\}$, where $(1,0)$ indicates $\sigma_0 \succ \sigma_1$ while $(0,1)$ indicates $\sigma_1 \succ \sigma_0$. Here $\sigma_i \succ \sigma_j$ implies segment $i$ is preferred to segment $j$. The probability of one segment being preferred over another is modeled via the Bradley-Terry model~\cite{bradley1952rank}:
\begin{equation}
\label{eq:btmodel}
    P_{\theta}[\sigma_1 \succ \sigma_0] = \frac{\exp{\sum_{t}} r_{\theta}(s_t^1, a_t^1)}{\sum_{i\in\{0,1\}}\exp{\sum_{t}} r_{\theta}(s_t^i, a_t^i)} \ .
\end{equation}
To train the reward function $r_\theta$, we minimize the cross-entropy loss between the predicted preferences $P_{\theta}$ and the observed preferences $y$,
\begin{equation}\label{eq:btloss}
\begin{split}
    \mathcal{L}_{\text{pref}} = - \mathbb{E}_{(\sigma_0,\sigma_1,y)\sim D} \big[&y(0)P_\theta[\sigma_0 \succ\sigma_1] + \\ &y(1)P_\theta[\sigma_1\succ\sigma_0]\big] \ .
\end{split}
\end{equation}
In preference-based RL, a policy $\pi_\phi$ and the reward
function $r_\theta$ are updated in an alternating fashion. First, the reward function is updated using the sampled preferences as described above. Next, the policy is optimized to maximize the expected cumulative reward under the learned reward function using standard RL algorithms. 

\begin{figure*}[!t]
    \centering
    \includegraphics[width=\textwidth]{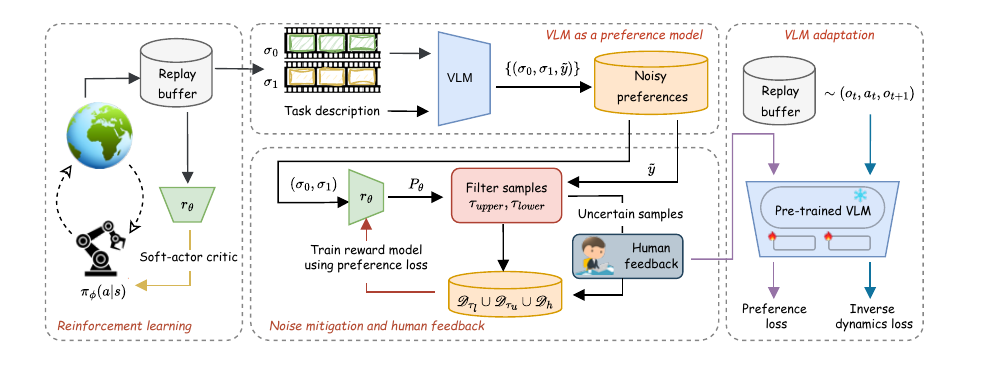}
    \caption{Overview of our approach. Given a task description, \algo~iteratively updates the policy $\pi_\phi$ via reinforcement learning using the reward model $r_\theta$. Trajectory segments from the replay buffer are sampled and labeled with VLM-generated preferences. These samples are then classified as clean or noisy using thresholds $\tau_{upper}$ and $\tau_{lower}$. A budgeted subset of noisy samples is sent for human annotation. The reward model is trained on both VLM and human-labeled preferences, while the VLM is fine-tuned using human annotations and replay buffer samples. 
    }
    \label{fig:overview}
\end{figure*}
\subsection{VLM as a reward model} \label{sec:vlm}
The use of VLMs in RL has primarily centered around CLIP-style models~\cite{radford2021learning}. These models comprise a language encoder $\mathcal{F}_L$ and an image encoder $\mathcal{F}_I$, which map text and image inputs into a shared latent space. Through contrastive learning on image-caption pairs, often augmented with task-specific or dynamics-aware objectives~\cite{ma2023liv}, VLMs align textual and visual representations effectively.

Given the image observation $o_t$ corresponding to a state $s_t$, and the language description of the task $l$, most works~\cite{mahmoudieh2022zero,adeniji2023language,fu2024furl} define the reward as:
\begin{equation}
\label{eq:cosine_sim}
    r^{vlm}_t = \frac{\langle \mathcal{F}_{L}(l), \mathcal{F}_{I}(o_t) \rangle}{\Vert \mathcal{F}_{L}(l) \Vert \cdot \Vert \mathcal{F}_{I}(o_t) \Vert} \ .
\end{equation}
However, this reward is often too coarse for fine-grained tasks like manipulation. Figure~\ref{fig:noisy_reward} shows the VLM reward curve for an expert trajectory, computed using~\eqref{eq:cosine_sim}. Ideally, the curve should align with the expert’s progress, assigning higher rewards as the state nears task completion. While the curve captures some aspects of task progress, it fails to fully reflect the expert’s trajectory. Given these limitations, fine-tuning VLMs with human feedback becomes essential. 

\section{Method} \label{sec:method}
In this section, we present \algo, a framework designed to minimize the reliance on time-intensive human supervision in preference-based RL by leveraging VLMs. 
We first outline how VLMs are utilized to provide preference feedback for training a reward model (Sec.~\ref{sec:method_vlm_as_pref}). Then, we address the limitations of directly applying VLMs to new environments and propose a data-efficient adaptation approach that combines self-supervised learning with minimal human feedback to align the VLM with the environment's dynamics (Sec.~\ref{sec:method_vlm_adapt}). 
Finally, to ensure robust training in the presence of noisy machine-generated feedback, we introduce a noise-mitigation mechanism that selectively identifies unreliable VLM outputs and refines them with minimal human input, achieving more stable and accurate reward model learning (Sec.~\ref{sec:method_sample_select}). 
Figure~\ref{fig:overview} provides an overview of our approach.

\subsection{VLM as a preference feedback model} \label{sec:method_vlm_as_pref}
To leverage VLMs for generating preference feedback, we begin by extracting image sequences corresponding to each trajectory segment. 
Specifically, for a given pair of segments $(\sigma_0, \sigma_1)$, we extract the image sequences $(O_0, O_1)$, where each sequence is defined as $O_i = \{o^i_0, o^i_1, o^i_2, \dots, o^i_T\}$ for $i \in \{0, 1\}$. 
Here, $o_t$ represents the image observation of the state $s_t$ at the $t$-th time step.

Using the language description of the task $l$, we compute the return for each segment as $R_i = \sum_{t=0}^T r^{vlm}_t$, with $r^{vlm}_t$ the reward at time step $t$ derived from the VLM via~\eqref{eq:cosine_sim}. 
Based on these returns, the preference label $\tilde{y}$ is assigned as: 
\begin{equation} \label{eq:vlm_pref} 
    \tilde{y} = \begin{cases} 
        (0, 1), & \text{if } R_0 > R_1, \\ 
        (1, 0), & \text{otherwise}. 
    \end{cases} 
\end{equation}
These preferences are then used to train a reward model $r_\theta(s_t, a_t)$ by minimizing the preference loss in~\eqref{eq:btloss}. 
The trained reward model can be integrated into a preference-based RL algorithm for policy optimization. 
In this work, we specifically leverage PEBBLE~\cite{lee2021pebble}, a preference-based RL framework that combines unsupervised pre-training with off-policy learning using Soft Actor-Critic (SAC)~\cite{haarnoja2018soft}.

\subsection{VLM adaptation} \label{sec:method_vlm_adapt}
A key challenge in directly applying VLMs to downstream RL tasks is the domain gap between their pretraining data and the target environment~\cite{raychaudhuri2021cross,fu2024furl}. 
This mismatch often results in noisy or misaligned feedback. 
To address this, we propose a data-efficient adaptation strategy that combines minimal human feedback with self-supervised learning to align the VLM with the target environment's dynamics.

We freeze the VLM and introduce two learnable layers, $\mathcal{G}_L$ and $\mathcal{G}_I$, on top of the language and image embedding layers, respectively. 
These layers are fine-tuned to adapt the VLM. 
The layer $\mathcal{G}_L$ processes the language embedding $\mathcal{F}_L(l)$ of the task description $l$ and produces an adapted text embedding, $\mathcal{G}_L\circ\mathcal{F}_L(l)$. 
Similarly, for each image observation $o_t$, the adapted image embedding is generated by $\mathcal{G}_I\circ\mathcal{F}_I(o_t)$. 
Using these adapted embeddings, the reward function for preference feedback is updated as: 
\begin{equation} \label{eq:cosine_sim_aligned} 
    r^{vlm}_t = \frac{\langle \mathcal{G}_L\circ\mathcal{F}_L(l), \mathcal{G}_I\circ\mathcal{F}_I(o_t) \rangle}{\Vert \mathcal{G}_L\circ\mathcal{F}_L(l) \Vert \cdot \Vert \mathcal{G}_I\circ\mathcal{F}_I(o_t) \Vert}. 
\end{equation}
A small number of human-provided preferences are sampled to fine-tune the VLM using the preference loss~\eqref{eq:btloss}. 
The dense rewards for training are derived from the updated similarity measure~\eqref{eq:cosine_sim_aligned}. 
The methodology for selecting human feedback samples is detailed in Section~\ref{sec:method_sample_select}.

We also fine-tune the VLM using an unsupervised objective designed to align the VLM embeddings with environment dynamics. 
Given the current observation $o_t$, action $a_t$, and the next observation $o_{t+1}$, we train the VLM to predict the action which leads to the transition between observations: 
\begin{equation} \label{eq:inv_loss} 
    \Vert f\left(\mathcal{G}_I\circ\mathcal{F}_I(o_t), \mathcal{G}_I\circ\mathcal{F}_I(o_{t+1})\right) - a_t \Vert^2, 
\end{equation} 
where $f$ is a linear layer. 
This encourages the adapted embeddings to capture task-relevant dynamics, improving the precision of preference feedback.

\subsection{Noise mitigation and human feedback} \label{sec:method_sample_select}
Machine-generated preferences, while scalable, are prone to noise and lack the reliability afforded by human-provided annotations. 
To ensure robust training, it is thus crucial to distinguish between accurate and noisy samples. 
This categorization not only improves the stability of the reward model training but also optimizes the use of human feedback by focusing on refining the noisy samples. 

\noindent \textbf{Identifying noisy samples.} Our approach is grounded in insights from research on noisy training~\cite{li2020gradient,cheng2024rime}, which shows that deep neural networks tend to learn generalizable patterns in the early stages of training before overfitting to noise in the data. Leveraging this observation, we prioritize samples with lower training losses as clean, while treating high-loss samples as potential candidates for human review. 

To formalize this, we use the preference loss defined in~\eqref{eq:btloss} to train the reward model $r_\theta$. 
Assuming the loss for clean samples is bounded by $\rho$, \citet{cheng2024rime} have shown that the KL divergence between the predicted preference distribution $P_{\theta}$ and the true preference label $\tilde{y}$ for a sample $(\sigma_0, \sigma_1)$ is lower-bounded, that is:
\begin{equation}\label{eq:KL_div} 
    D_{KL}(\tilde{y}\Vert P_\theta) \geq -\ln{\rho} + \frac{\rho}{2} + \mathcal{O}(\rho^2) \ . \end{equation}
To filter out unreliable samples, we take a lower bound on the KL divergence, $\tau_{base} = -\ln{\rho} + \alpha \rho$, where $\rho$ is the maximum loss calculated on the filtered samples from the latest update, and $\alpha$ is a tunable hyperparameter, with a theoretical range of $(0, 0.5]$. 
However, because of shifts in the state distribution during RL training, we employ another auxiliary term $\tau_{unc} = \beta_{t} \cdot s_{KL}$, to account for uncertainty. 
Here, $\beta_{t}$ is a time-dependent parameter, and $s_{KL}$ is the standard deviation of KL divergence. 
The corrected lower bound is given by $\tau_{lower} = \tau_{base} + \tau_{unc}$. 
To control the uncertainty over time, $\beta_t$ follows a linear decay schedule that allows greater tolerance for noisy samples in the early stages of training while becoming more conservative as training progresses. 
Specifically, $\beta_t = \max(\beta_{min}, \beta_{max} - kt)$, where $\beta_{min}$ and $\beta_{max}$ are fixed values, and $k$ is a constant that controls the rate of decay. 

\noindent \textbf{Handling noisy samples.} Samples with a KL divergence below $\tau_{lower}$ are considered clean and are incorporated into the reward model training: 
\begin{equation} 
\label{D_lower}
    \mathcal{D}_{\tau_{l}} = \{(\sigma^0, \sigma^1, \tilde{y}) : D_{KL}(\tilde{y} \Vert P_\theta(\sigma_0, \sigma_1)) < \tau_{lower}\} \ . 
\end{equation}
Conversely, samples with a KL divergence exceeding a higher threshold $\tau_{upper}$ are presumed to contain noisy labels. To preserve their utility, we relabel these samples by flipping their predicted labels and include them in a separate dataset:
\begin{equation}
\label{D_upper}
    \mathcal{D}_{\tau_u} = \{(\sigma_0, \sigma_1, \mathbf{1}-\tilde{y}) : D_{KL}(\tilde{y} \Vert P_\theta(\sigma_0, \sigma_1)) > \tau_{upper}\} \ .
\end{equation}
The remaining samples, with KL divergence between $\tau_{lower}$ and $\tau_{upper}$, are deemed uncertain, and we randomly sample from them based on the available human annotation budget. 
These samples are particularly valuable, as both the VLM and reward model struggle to assign reliable labels. 
By focusing human effort on this subset, $\mathcal{D}_{h}$, we ensure that annotations address the most challenging cases. 

\noindent \textbf{Training with selective feedback.} The reward model is trained on $N_{\text{vlm}} = |\mathcal{D}_{\tau_l}| + |\mathcal{D}_{\tau_u}|$ machine-labeled samples, supplemented by $N_{\text{human}}=|\mathcal{D}_h|$ human-labeled samples from uncertain cases. 
Only $N_{\text{human}}$ samples are used to update the VLM. 
This targeted feedback mechanism accelerates training convergence and improves accuracy while significantly reducing the overall annotation burden. 

\subsection{Overall algorithm} \label{sec:method_overall}
\algo~proceeds by initializing the policy $\pi_\phi$, reward function $r_\theta$, and additional layers $\mathcal{G}$ on top of the VLM randomly. 
Given a task description $l$, the method iteratively follows a structured cycle. 
First, the policy $\pi_\phi$ is updated using 
the reward function $r_\theta$, interacting with the environment and storing observed trajectories in a buffer. From this buffer, trajectory segment pairs are randomly sampled and assigned preferences using the VLM. 
These labeled pairs are then categorized into clean and noisy samples based on the filtering strategy outlined in Section~\ref{sec:method_sample_select}. 
A subset of the noisy samples is sent for human annotation, subject to a predefined budget. 
The reward model is updated using the preference-labeled pairs (both VLM and human annotated) through~\eqref{eq:btloss}, while the VLM is fine-tuned using the human-annotated samples, following the adaptation strategy outlined in Section~\ref{sec:method_vlm_adapt}. 
The pseudo-code of our approach is shown in Algorithm~\ref{alg:prefvlm} in the appendix.

%% file: sections/4_experiments.tex
\begin{figure}[t]
    \centering
    \includegraphics[width=\columnwidth]{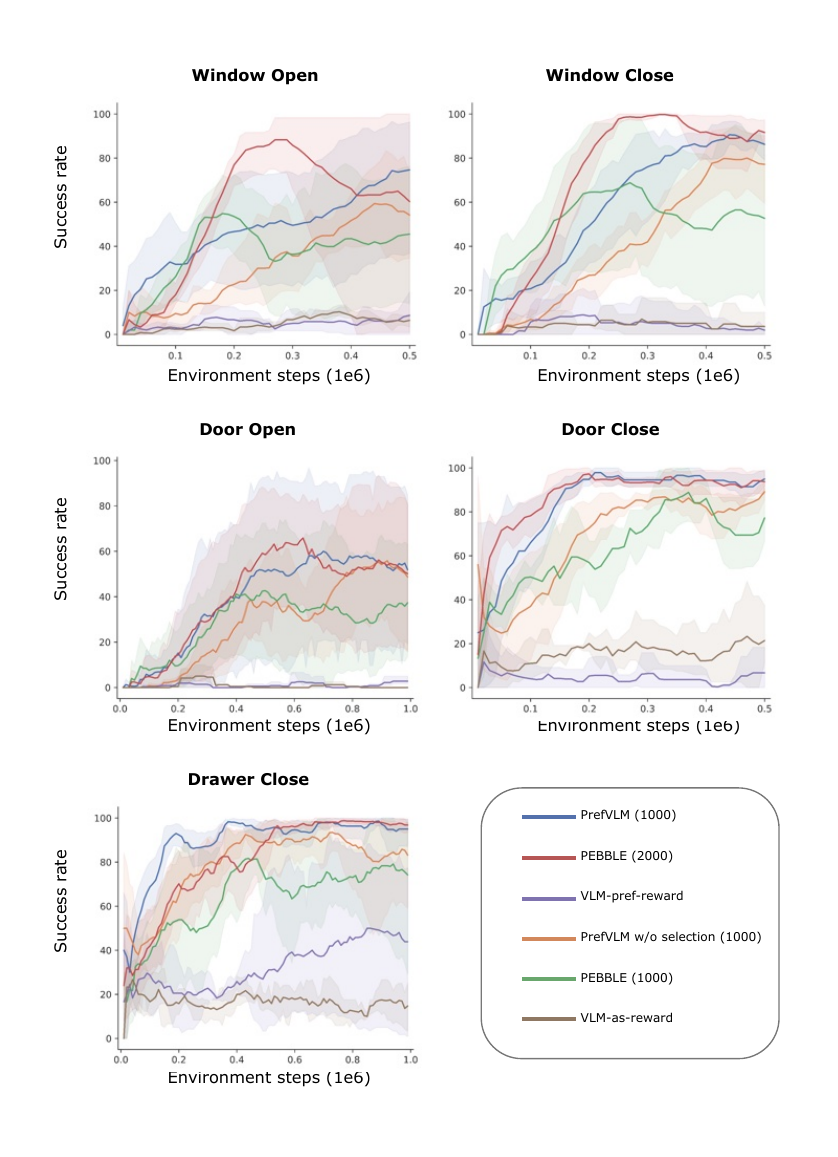}
    \caption{Learning curves for all methods on the 5 Meta-World tasks. \algo~consistently outperforms all baselines with minimal human feedback and matches or exceeds PEBBLE’s performance while using $2\times$ fewer annotations. Results are averaged over 5 seeds, with shaded regions indicating the standard error. 
    }
    \label{fig:exp_main}
\end{figure}
\section{Experiments} \label{sec:experiments}

To assess the effectiveness of \algo~in leveraging the synergy between VLMs and human feedback to improve the efficiency of preference-based RL we seek to answer:

$\bullet$ Can \algo~achieve comparable task success with significantly less human feedback than standard preference-based RL methods?

$\bullet$ Given the same amount of human input, does \algo~yield higher task success than existing methods?

$\bullet$ Can \algo~transfer its knowledge from one task to similar tasks, further reducing the need for human feedback?

\subsection{Experimental setup} \label{sec:exp_setup}
We evaluate \algo~on five manipulation tasks from Meta-World~\cite{yu2020meta}: \emph{door open}, \emph{drawer close}, \emph{drawer open}, \emph{window open}, and \emph{window close}. 
SAC~\cite{haarnoja2018soft} is used as the RL policy learning algorithm, while the VLM is initialized with the pre-trained LIV~\cite{ma2023liv} model. 
We optimize using Adam~\cite{kingma2014adam} with a learning rate of 0.0001. 
reference feedback is provided for trajectory segments of length 50. 
The policy is learned with state observations. 
Additional details on implementation and environments are provided in Appendix~\ref{app:env},~\ref{app:impl}.

\subsection{Baselines} \label{sec:exp_baseline}
We compare \algo~against the following baselines:
\begin{enumerate}
    \item \textbf{VLM-as-reward}~\cite{rocamonde2023vision}: This approach directly uses a VLM to assign rewards to each state following~\eqref{eq:cosine_sim}.
    \item \textbf{PEBBLE}~\cite{lee2021pebble}: A preference-based RL method which uses human feedback to learn policies using SAC.
    \item \textbf{VLM-pref-reward}: A modification of the above where segment pairs are ranked based on~\eqref{eq:vlm_pref} instead of a human. 
    These annotated segment pairs are used to train the reward function $r_{\theta}$ in PEBBLE.
    \item \textbf{\algo~w/o selection}: A variant of our method that does not exploit the noise mitigation strategies crucial to human-machine interaction. 
    Here, human feedback is obtained for randomly selecting pairs of segments from the replay buffer. 
\end{enumerate}

\subsection{Main results} \label{sec:exp_results}

\paragraph{Does \algo~improve feedback efficiency?}
Figure~\ref{fig:exp_main} shows the learning curves for all methods, comparing task success rates of the learned policies. 
We evaluate \algo~with 1000 human feedback samples, while PEBBLE, the baseline preference-based method, is assessed with both 1000 and 2000 feedback samples to highlight the impact of feedback efficiency. 

Across all tasks, \algo~matches PEBBLE's performance while requiring only half the human feedback, demonstrating the effectiveness of VLMs in reducing human annotation costs without compromising learning performance. 
However, directly using VLMs as standalone reward models is ineffective, as both \textit{VLM-as-reward} and \textit{VLM-as-pref} fail to achieve meaningful success. 
This is due to the inherently noisy VLM outputs when applied without supervision (see Section~\ref{sec:vlm}), underscoring the necessity of integrating human feedback rather than relying on VLMs in a zero-shot manner. 

Additionally, while \textit{\algo~w/o selection} achieves reasonable performance and even surpasses PEBBLE with the same amount of feedback, it remains inferior to the full method. This is because it allows some noisy labels to influence reward model training and introduces redundancy in human annotations, as humans may not always label the most uncertain samples for the VLM. This further highlights the importance of selective feedback in optimizing human effort and maintaining robust learning.

\begin{figure}[t]
    \centering
    \includegraphics[width=\columnwidth]{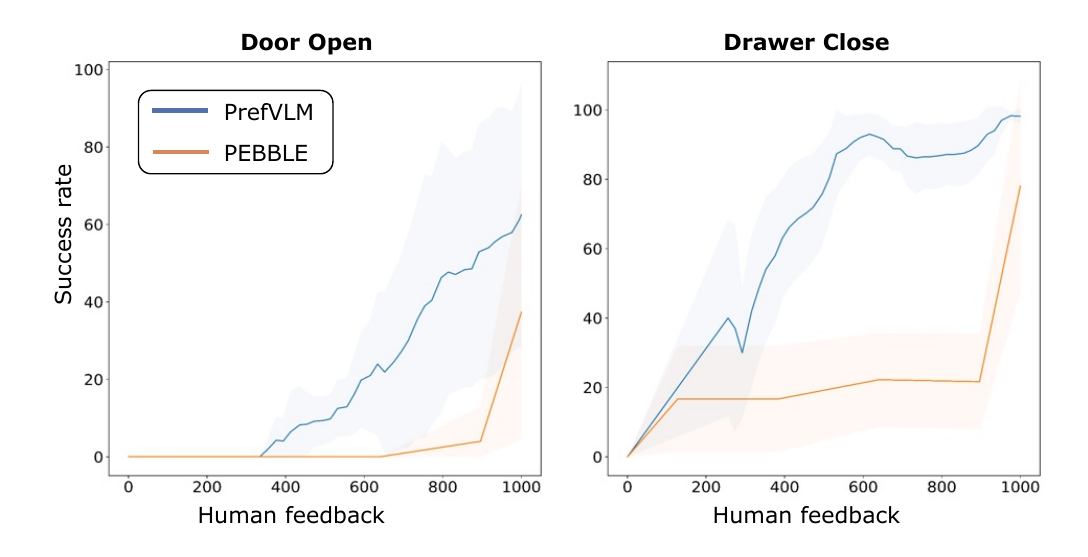}
    \caption{Success rate as a function of human feedback. \algo~leverages human feedback more efficiently by complementing it with VLM-based feedback, resulting in higher success rates with fewer human annotations. Results are averaged over 5 seeds, with shaded regions representing the standard error.}
    \label{fig:human_feedback}
\end{figure}

\paragraph{How does success rate improve with human feedback?} 
Since \algo~significantly reduces the need for human feedback in preference-based RL, we further analyze how task success rates scale with varying amounts of feedback. As shown in Figure~\ref{fig:human_feedback}, our approach consistently achieves higher success rates as the amount of human feedback increases. In contrast, PEBBLE shows slower growth in success rate with additional feedback, as it lacks the ability to leverage VLM-based feedback, unlike our method.

\paragraph{Can \algo~transfer knowledge across tasks?}
A key objective of \algo~is to refine the VLM with human feedback, mitigating its inherent noise. An important question is whether an already adapted VLM can generalize to related tasks with minimal additional supervision.

\begin{figure}[t]
    \centering
    \includegraphics[width=\columnwidth]{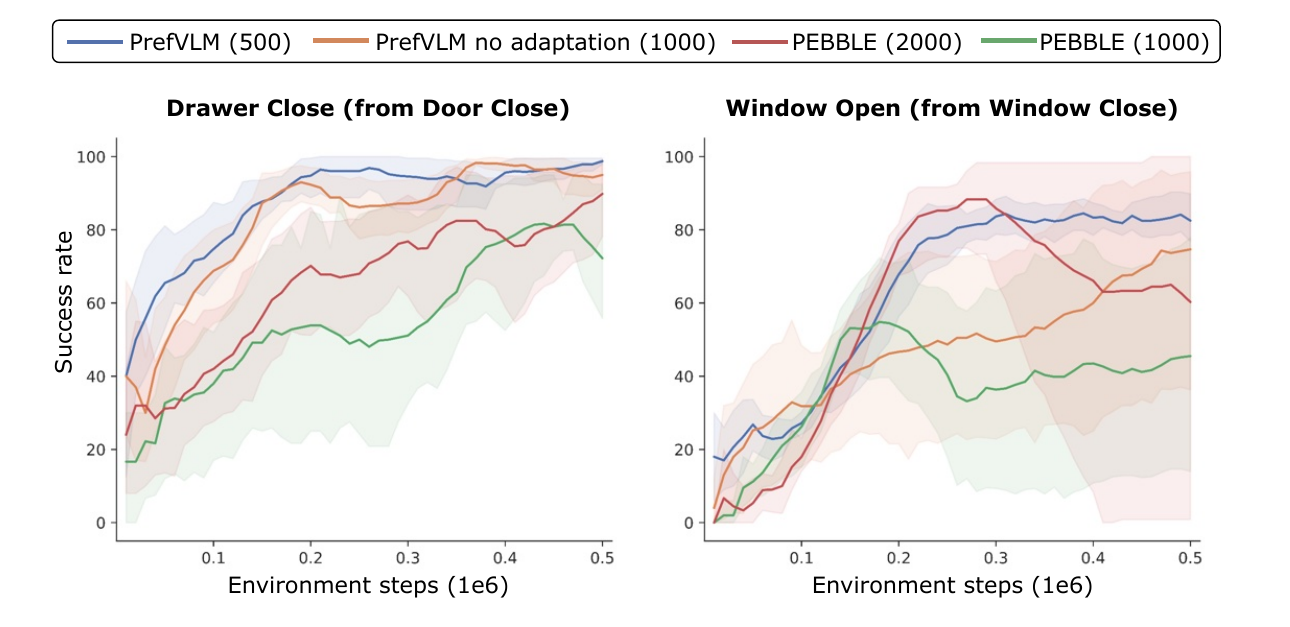}
    \caption{Learning curves for knowledge transfer experiments. \algo~achieves comparable or superior performance to PEBBLE while requiring $4\times$ fewer annotations, demonstrating its ability to transfer knowledge across both \emph{same task, different object} (left) and \emph{same object, different task} (right) settings. This highlights an additional pathway for reducing human feedback in preference-based RL. Results are averaged over 5 seeds, with shaded regions representing standard error.}
    \label{fig:transfer_results}
\end{figure}

To evaluate this, we consider two types of task transfer: (1) \emph{same task, different object}: adapting from ``door close'' to ``drawer close'', and (2) \emph{same object, different task}: adapting from ``window close'' to ``window open''. In both cases, we initialize the VLM for the target task with the weights obtained after adaptation on the source task, while keeping the rest of the algorithm unchanged. As shown in Figure~\ref{fig:transfer_results}, \algo~achieves the success rate of PEBBLE trained with 2000 human feedback samples using only 500 samples - a $4\times$ reduction in annotation effort. This demonstrates that adapting the VLM on one task enables efficient generalization to related tasks, significantly improving feedback efficiency in preference-based RL.

\subsection{Analysis} \label{sec:exp_analysis}

\begin{figure}[t]
    \centering
    \includegraphics[width=\columnwidth]{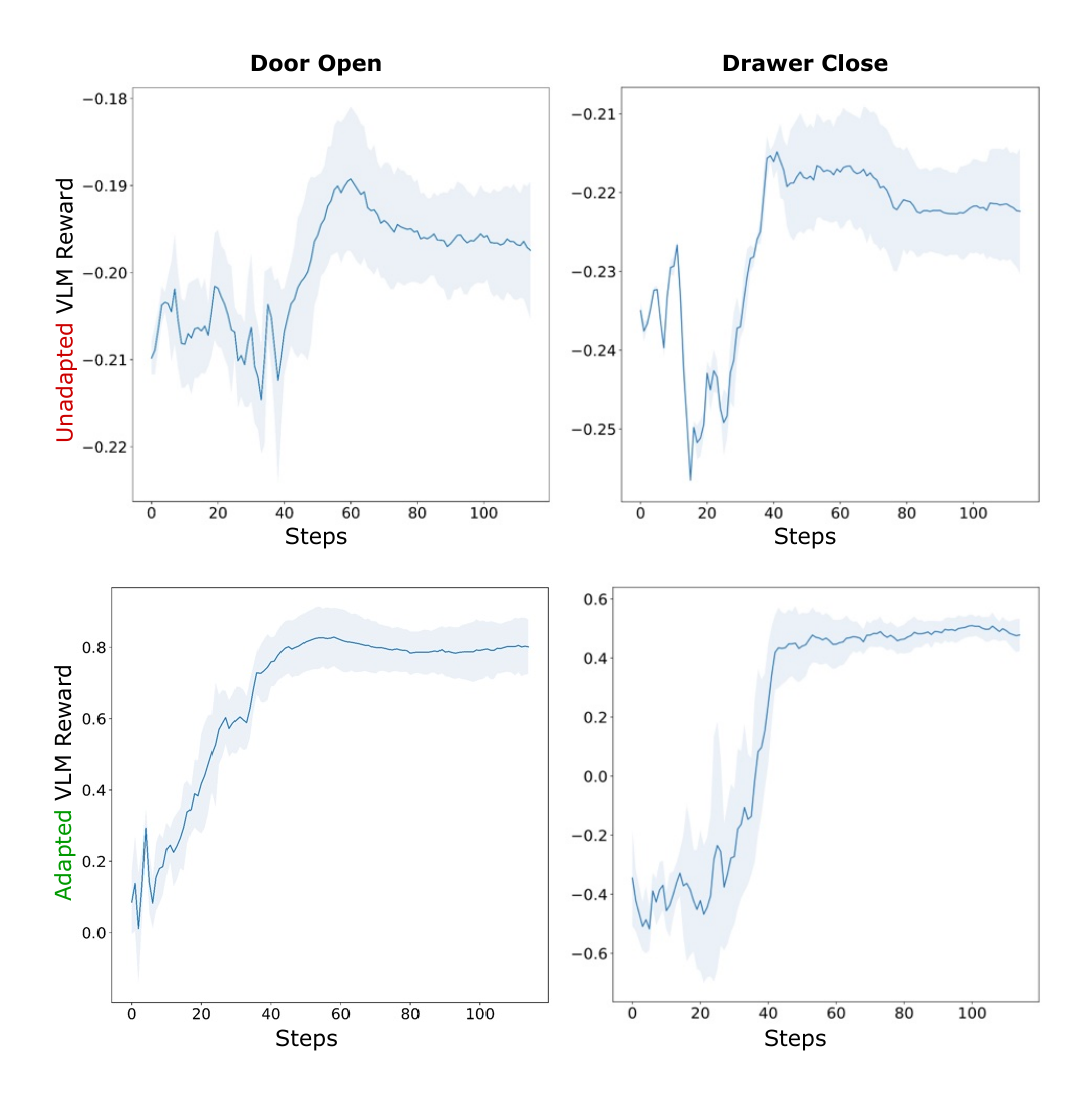}
    \caption{VLM reward on \emph{Door Open} and \emph{Drawer Close} before (top) and after (bottom) adaptation, averaged across the same five expert trajectories. The adapted VLM better aligns with ground-truth task progress.}
    \vspace{-0.2cm}
    \label{fig:adapted_vlm}
\end{figure}

\paragraph{Impact of VLM adaptation} 
Figure~\ref{fig:adapted_vlm} compares VLM rewards before and after adaptation, averaged over five expert trajectories. Ideally, rewards should increase along the trajectory, reflecting true task progress. As shown, the adapted VLM better aligns with ground-truth task progress, yielding higher reward values toward the end of trajectories when the task is successfully completed, without abrupt drops. This adaptation significantly reduces reliance on human feedback, allowing VLMs to serve as a more effective prior, mitigating the need to learn a reward model from scratch.

\begin{figure}[t]
    \centering
    \includegraphics[width=\columnwidth]{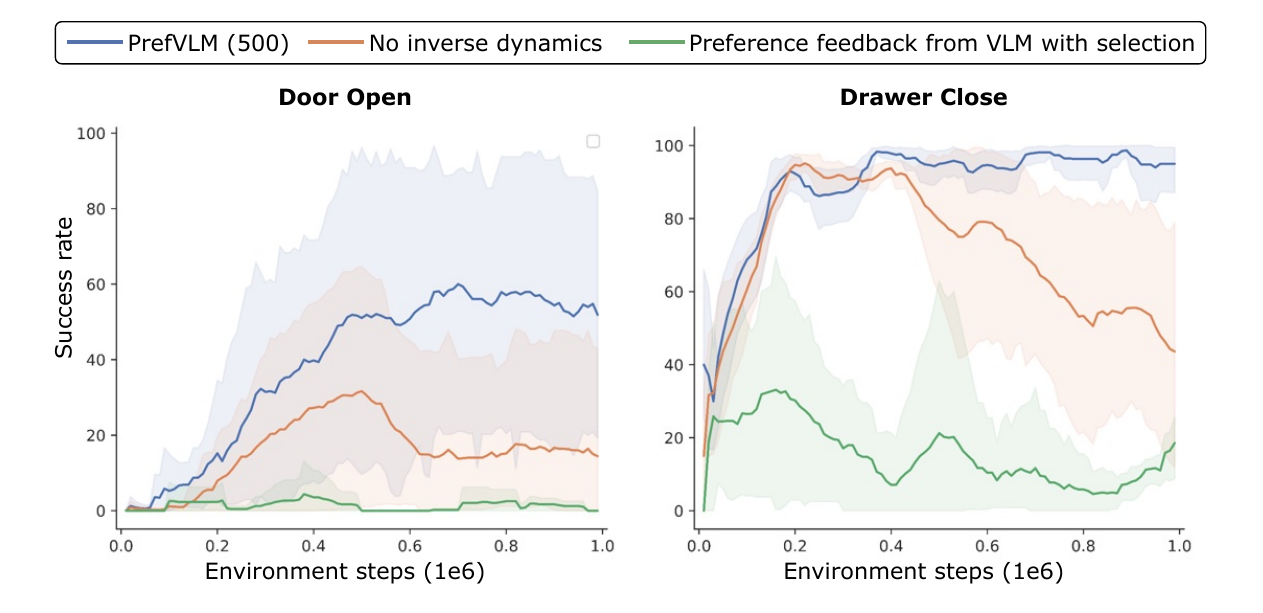}
    \caption{Success rates for (i) VLM adaptation without inverse dynamics loss and (ii) VLM-generated preferences with noise filtering. Results show that inverse dynamics loss stabilizes learning while filtering alone is ineffective without initial human feedback. Results are averaged over 5 seeds, with shaded regions representing standard error.}
    \vspace{-0.2cm}
    \label{fig:ablation}
\end{figure}

\paragraph{Impact of inverse dynamics loss.}  
To evaluate the efficacy of the self-supervised loss, we trained the VLM using only contrastive loss in~\eqref{eq:cosine_sim_aligned}. 
As shown in Figure~\ref{fig:ablation}, the success rate initially increases but then degrades after a certain number of steps. 
In the early stages, when human feedback is available, the VLM performs well by leveraging reliable supervision. However, without the inverse dynamics loss, its performance declines as the policy evolves. This degradation occurs because while the VLM learns from human preferences, these preferences are based on samples from a suboptimal policy. As the policy improves and the data distribution shifts, the VLM struggles to generalize due to its limited understanding of environment dynamics. In contrast, incorporating the inverse dynamics loss allows the VLM to adapt to distribution shifts, maintaining stable performance throughout training.

\paragraph{Preference feedback from VLM with noise mitigation.} \label{sec:exp_vlm_selection} 
We investigate whether preference feedback from VLMs (denoted as \emph{VLM-pref-reward} in Section~\ref{sec:exp_baseline}) can be effectively combined with the filtering strategy described in Section~\ref{sec:method_sample_select} to achieve success rates comparable to human feedback. As shown in Figure~\ref{fig:ablation}, this approach alone fails to yield significant improvements. The primary limitation arises from the inherent noise in VLM-generated preferences, leading to unreliable feedback. When the reward model is trained on these noisy labels, it learns a suboptimal reward function, and filtering samples based on KL divergence in cross-entropy loss further amplifies these errors, resulting in poor overall performance. This underscores the need for at least some human feedback to adapt the VLM. 

%% file: sections/5_conclusion.tex
\section{Conclusion}
We introduce a novel approach, \algo, to reduce human feedback requirements in preference-based reinforcement learning by leveraging vision-language models. While VLMs encode rich world knowledge, their direct application as reward models is hindered by alignment issues and noisy predictions. To address this, we develop a synergistic framework where limited human feedback is used to adapt VLMs, improving their reliability in preference labeling. Further, we incorporate a selective sampling strategy to mitigate noise and prioritize informative human annotations.

Our experiments demonstrate that this method significantly improves feedback efficiency, achieving comparable or superior task performance with up to 50\% fewer human annotations. Moreover, we show that an adapted VLM can generalize across similar tasks, further reducing the need for new human feedback by 75\%. These results highlight the potential of integrating VLMs into preference-based RL, offering a scalable solution to reducing human supervision while maintaining high task success rates. 

\section*{Impact Statement}
This work advances embodied AI by significantly reducing the human feedback required for training agents. This reduction is particularly valuable in robotic applications where obtaining human demonstrations and feedback is challenging or impractical, such as assistive robotic arms for individuals with mobility impairments. By minimizing the feedback requirements, our approach enables users to more efficiently customize and teach new skills to robotic agents based on their specific needs and preferences. The broader impact of this work extends to healthcare, assistive technology, and human-robot interaction. One possible risk is that the bias from human feedback can propagate to the VLM and subsequently to the policy. This can be mitigated by personalization of agents in case of household application or standardization of feedback for industrial applications.

%% file: sections/6_appendix.tex
\newpage
\appendix
\onecolumn
\section{Pseudo-code}
\label{app:pseudocode}
\begin{algorithm}[ht]
\caption{Language Guided RL through preference feedback from human and VLM (\algo)}\label{alg:guiderl}
\begin{algorithmic}[1]
\REQUIRE Text description of the task \textit{l}, maximum human feedback $M_{human}$, pre-trained VLM $\mathcal{F}$
\STATE Initialize policy $\pi_\phi$, reward model $r_\theta$, and VLM trainable layers $\mathcal{G}$
\STATE Initialize replay buffer $\mathcal{R}$, preference buffer $\mathcal{B}$, human preference buffer $\mathcal{H}$, feedback batch size $N$, and feedback collected counter $m_h$
\FOR{each iteration $i$}
\STATE Collect sample state $s_{i+1}$, image $o_{i}$ by taking $a_i \sim \pi_\theta(a_i|s_i)$
\STATE Add sample $\mathcal{R} \longleftarrow \mathcal{R} \cup \{s_i, a_i, s_{i+1}, r_\theta(s_i, a_i), o_i\}$
\FOR{each policy update step}
\STATE Sample  minibatch $\{(s_i, a_i, s_{i+1}, r_\theta(s_i, a_i))\} \sim \mathcal{R}$
\STATE Optimize policy $\pi_\phi$ using the sampled batch with any off-policy RL algorithm 
\ENDFOR
\IF{$i$ mod $K == 0$}
\FOR{$n = 1$ to $N$}
\STATE Sample preference pair $(\sigma_0, \sigma_1)\sim\mathcal{R}$
\STATE Obtain preference feedback $\tilde{y}$ from the VLM
\STATE Update $\mathcal{B} \longleftarrow \mathcal{B} \cup \{\sigma_0, \sigma_1, \tilde{y}\}$
\ENDFOR
\IF{$m_{h} < M_{human}$}
\STATE Obtain $\mathcal{D}_{\tau_l}$ and $\mathcal{D}_{\tau_u}$ from $\mathcal{B}$ using Eqn.~(\ref{D_lower}) and Eqn.~(\ref{D_upper}).
\STATE Sample $\mathcal{D}_h$ from the $\mathcal{B} - \mathcal{H}$, such that $|\mathcal{D}_h| = \min(|\mathcal{B}| - |\mathcal{D}_{\tau_l}| - |\mathcal{D}_{\tau_u}|, 0.05*N)$ 
\STATE Obtain human feedback $y$ for $\mathcal{D}_h$.
\STATE Update $\mathcal{H} \longleftarrow \mathcal{H} \cup \{\mathcal{D}_h, y\}$.
\STATE Update $\mathcal{B} \longleftarrow \mathcal{B} \cup \{\mathcal{D}_h, y\}$
\STATE $m_h = m_h + |\mathcal{D}_h|$
\FOR{each VLM update step}
\STATE Update the VLM layers $\mathcal{G}$ with $\mathcal{H}$ using Eqn~(\ref{eq:btloss}) and Eqn.~(\ref{eq:inv_loss})
\ENDFOR
\ENDIF
\FOR{each reward model update step}
\STATE Obtain $\mathcal{D}_{\tau_l}$ and $\mathcal{D}_{\tau_u}$ from $\mathcal{B}$ using Eqn.~(\ref{D_lower}) and Eqn.~(\ref{D_upper})
\STATE Optimize $r_\theta$ with $\mathcal{D}_{\tau_l}$, $\mathcal{D}_{\tau_u}$ and $\mathcal{H}$, with Eqn.~(\ref{eq:btloss}) 
\ENDFOR
\STATE Relabel $\mathcal{R}$ with $r_\theta$.
\ENDIF

\ENDFOR
\end{algorithmic}  
\label{alg:prefvlm}
\end{algorithm}

\section{Environmental details and task descriptions}
\label{app:env}
\begin{figure}[H]
    \centering
    \includegraphics[width=0.8\linewidth]{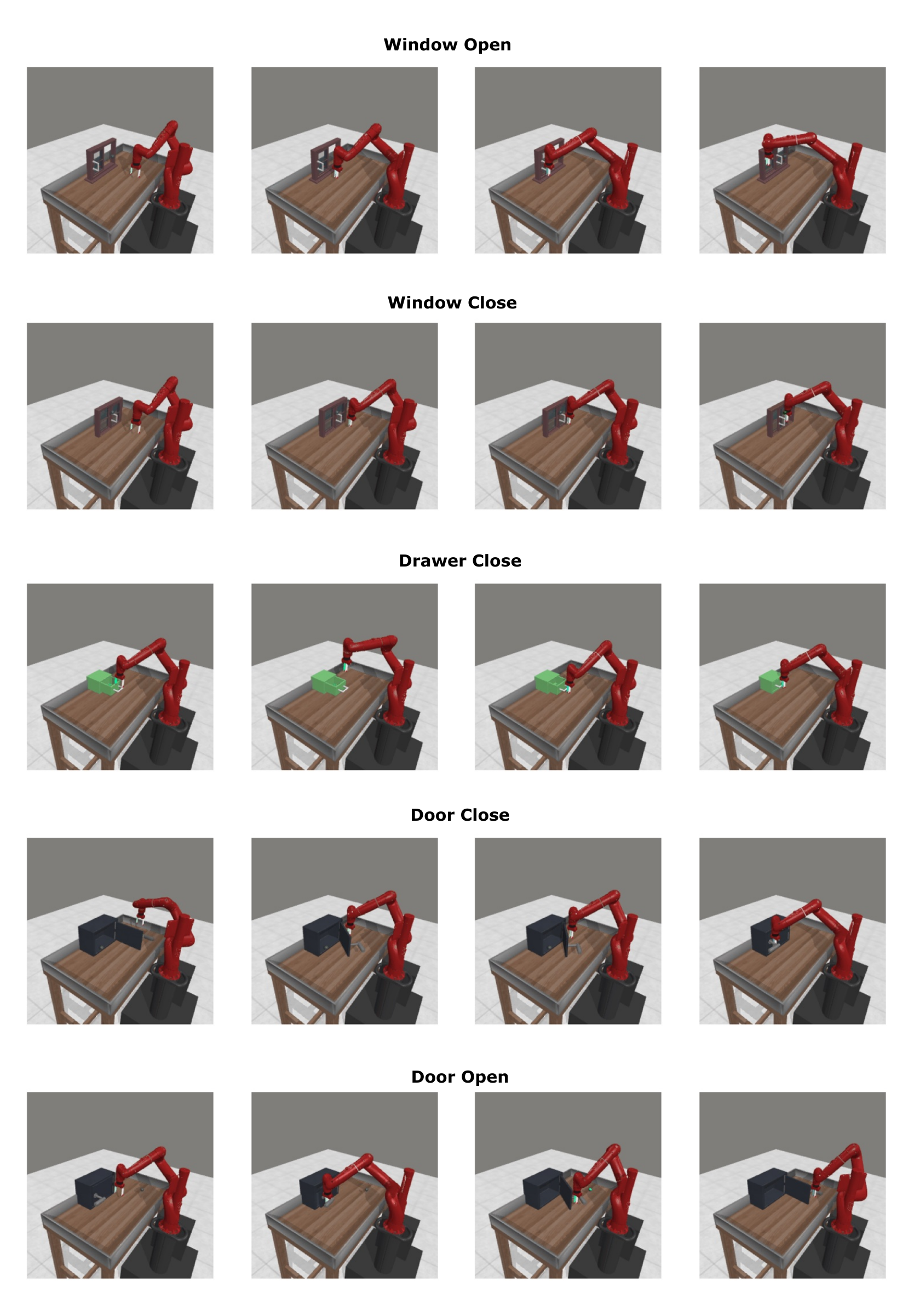}
    \caption{Example expert trajectories demonstrating the tasks we test on.}
    \label{fig:env}
\end{figure}

We show results on 5 different environments on Meta World as shown in Figure~\ref{fig:env}. For each of the environment, the state $s\in\mathbb{R}^{39}$, and the action $a\in\mathbb{R}^4$. Observations are captured from Camera 2 and rendered as 300 × 300 images. Task descriptions are sourced directly from the Meta-World paper~\cite{yu2020meta}. The corresponding prompts for each environment are provided below:
\begin{itemize}
    \item \textbf{door-open-v2} : \textit{Open a door with a revolving joint}
    \item \textbf{door-close-v2} : \textit{Push and close a door with a revolving joint}
    \item \textbf{window-open-v2} : \textit{Push and open a window}
    \item \textbf{window-close-v2} : \textit{Push and close a window}
    \item \textbf{drawer-close-v2} : \textit{Push and close a drawer}
\end{itemize}

\section{Implementation details}
\label{app:impl}
In our experiments, we begin by collecting initial data using a random policy for the first 1000 interactions with the environment. Following this, we conduct unsupervised exploration for the next 5000 interactions, during which we update state entropy for Soft Actor-Critic (SAC). Once 6000 interactions are completed, we initiate the training process for the policy, reward model, and VLM.

During the early training phase, we allocate 250 human feedback samples from the total budget ($M_{human}$) to train both the VLM and reward model, setting the initial feedback allocation as $m_h = 250$. After this initialization, training proceeds as outlined in Appendix~\ref{app:pseudocode}, following the full algorithmic procedure. We set $K=3000$. 

For the reward model, we use an ensemble of 3 MLPs, each with 3 hidden layers of 256 nodes and Leaky ReLU activation, while the final layer applies a tanh activation. The reward model is trained with a learning rate of 0.0003, a batch size of 128, and 200 update steps per iteration. The maximum feedback budget is set to 30000 samples (this includes the human feedback $M_{human}$ and the VLM preference feedback).

We use LIV~\cite{ma2023liv} as the VLM in our algorithm. The trainable layers $\mathcal{G}_L$ and $\mathcal{G}_I$ are both 2-layer MLPs with 256 and 64 hidden units respectively, with ReLU activation. The final layer does not have any activation. The inverse dynamics prediction layer $f$ is implemented as another MLP with 128 length input layer, hidden layer of size 64 with ReLU activation, and output layer of 4 units which is the dimension of the action vector. The learning rate is set to 0.0003. 

For the SAC policy, both the actor and critic are MLPs with three hidden layers of 256 units each, trained with a learning rate of 0.0001, with the critic $\tau$ set to the default value of 0.005.

Lastly, for label selection, we adopt the same hyperparameter values as reported in \cite{cheng2024rime}. Specifically, we set $\alpha = 0.5$, $\beta_{min} = 1$,  $\beta_{max} = 3$, $k = 1/300$ and $\tau_{upper} = 3\ln(10)$. 